\documentclass[journal]{IEEEtran}

\usepackage[T1]{fontenc}
\usepackage[utf8]{inputenc} 

\usepackage{amsmath,amssymb,amsfonts}
\usepackage{mathtools}
\usepackage{graphicx}

\usepackage{booktabs,makecell,multirow,tabularx}
\usepackage{array}
\usepackage{algorithm}
\usepackage{algorithmic} 

\usepackage{pifont}

\usepackage{hyperref}

\begin{document}

\title{Attention-Driven Framework for Non-Rigid Medical Image Registration}

\author{Muhammad Zafar Iqbal, Ghazanfar Farooq Siddiqui, Anwar Ul Haq, Imran Razzak,
\thanks{Muhammad Zafar Iqbal is with the Department of Computer Sciences, Quaid-i-Azam University (QAU), Islamabad, Pakistan.(e-mail: mziqbal@cs.qau.edu.pk).}
\thanks{Ghazanfar Farooq Siddiqui is with the Department of Computer Sciences, Quaid-i-Azam University (QAU), Islamabad, Pakistan.(e-mail: ghazanfar@qau.edu.pk).}

\thanks{Anwar Ul Haq is with the School of Engineering and Technology, Centre for Intelligent Systems, CQU Sydney, Australia (e-mail: a.anwaarulhaq@cqu.edu.au).}
\thanks{Imran Razzak is with MBZUAI, Abu Dhabi and  UNSW Sydney, NSW, Australia (e-mail: imran.razzak@mbzuai.ac.ae).}}

\maketitle
\begin{abstract}

Deformable medical image registration is a fundamental task in medical image analysis with applications in disease diagnosis, treatment planning, and image-guided interventions. Despite significant advances in deep learning based registration methods, accurately aligning images with large deformations while preserving anatomical plausibility remains a challenging task. In this paper, we propose a novel Attention-Driven Framework for Non-Rigid Medical Image Registration (AD-RegNet) that employs attention mechanisms to guide the registration process. Our approach combines a 3D UNet backbone with bidirectional cross-attention, which establishes correspondences between moving and fixed images at multiple scales. We introduce a regional adaptive attention mechanism that focuses on anatomically relevant structures, along with a multi-resolution deformation field synthesis approach for accurate alignment. The method is evaluated on two distinct datasets: DIRLab for thoracic 4D CT scans and IXI for brain MRI scans, demonstrating its versatility across different anatomical structures and imaging modalities. Experimental results demonstrate that our approach achieves performance competitive with state-of-the-art methods on the IXI and DIRLab datasets. The proposed method maintains a favorable balance between registration accuracy and computational efficiency, making it suitable for clinical applications. A comprehensive evaluation using normalized cross-correlation (NCC), mean squared error (MSE), structural similarity (SSIM), Jacobian determinant, and target registration error (TRE) indicates that attention-guided registration improves alignment accuracy while ensuring anatomically plausible deformations.
\end{abstract}



\section{Introduction}
Medical image registration is the process of spatially aligning two or more images to establish anatomical or functional correspondences. It plays a crucial role in various clinical applications, including disease diagnosis, treatment planning, image-guided interventions, and longitudinal studies~\cite{ding2024c2fresmorph, iqbal2024hybrid, sotiras2013deformable}. Among different registration types, deformable registration, which estimates a dense displacement field to align images via non-linear transformations, is vital for handling complex anatomical variations and organ motion~\cite{chen2022transmorph}. Traditional deformable registration methods typically formulate the problem as an optimization task, where a similarity metric between the fixed and moving images is maximized while regularization terms ensure smoothness and plausibility of the deformation~\cite{chen2021vit, avants2011reproducible}. However, these methods often suffer from high computational complexity, sensitivity to parameter settings, and limited ability to capture large deformations~\cite{ding2024c2fresmorph}.

In recent years, deep learning approaches have revolutionized medical image registration by employing the representational power of neural networks to learn the complex mapping between image pairs~\cite{balakrishnan2019voxelmorph, de2019deep, cao2024light, chen2024survey, liu2025focusmorph}. These methods are commonly grouped into supervised (trained with ground-truth deformation fields), unsupervised (optimize image similarity without labels), and weakly/semi supervised approaches (use sparse annotations such as landmarks, segmentations, or correspondence hints). While deep learning-based methods have significantly improved registration speed and accuracy, they still face challenges in handling large deformations, preserving anatomical plausibility, and focusing on clinically relevant structures~\cite{liu2025focusmorph, chen2024survey}.

Attention mechanisms have emerged as powerful tools across various computer vision tasks, enabling models to focus on salient features while suppressing irrelevant information~\cite{vaswani2017attention}. In medical image registration, attention can help identify corresponding anatomical structures and guide the deformation process more effectively. Early work introduced self-attention into registration pipelines~\cite{chen2021vit, chen2022transmorph}, with promising results. More recent advancements employ cross-attention between moving and fixed images to model spatial correspondences explicitly~\cite{chen2023deformable, meng2024advancing, wang2024pyramid}. Hierarchical attention designs and nested attention fusion strategies have also demonstrated improved performance in capturing both global and local deformations~\cite{kumar2025nestedmorph}.

Despite these advancements, challenges remain in achieving interpretable and anatomically consistent registration, particularly for large-deformation cases and across different modalities. In this paper, we propose a novel Attention-Driven Framework for Non-Rigid Medical Image Registration (AD-RegNet) that employs attention mechanisms to guide the registration process. Our approach makes the following contributions.

\begin{enumerate}
    \item We introduce a bidirectional cross-attention module that explicitly models the relationships between features from moving and fixed images at multiple scales, enabling more accurate correspondence establishment.
    
    \item We propose a regional adaptive attention mechanism that identifies and focuses on anatomically relevant structures, improving registration accuracy in regions of interest.
    
    \item We conduct comprehensive experiments on two distinct datasets (DIR-Lab for thoracic CT and IXI for brain MRI), demonstrating the versatility and effectiveness of our approach across different anatomical structures and imaging modalities.
\end{enumerate}

\section{Related Work}
\label{sec:related_work}
\subsection{Traditional Deformable Image Registration}
Conventional deformable image registration (DIR) methods have been extensively studied and are grounded in physical models and optimization theory. Elastic-model-based approaches, such as B-spline-based techniques~\cite{jiang2018cnn}, capture local deformations by modeling relationships between control points and spline functions. Fluid-based models employ viscous flow formulations to handle large deformations, enhancing registration accuracy in dynamic structures~\cite{beg2005computing}. Symmetric diffeomorphic methods, such as SyN~\cite{avants2008symmetric}, introduced symmetric energy formulations and normalized cross-correlation (NCC) to ensure topology-preserving and invertible deformations. Optical flow-based techniques reformulate registration as the estimation of a velocity field~\cite{vercauteren2009diffeomorphic}. While these methods achieve high accuracy, they rely on iterative optimization in high-dimensional parameter spaces, which can be computationally expensive and time-consuming, limiting their applicability in time-sensitive settings such as surgical guidance.

\begin{figure}[htbp]
    \centering
        \includegraphics[width=1\linewidth, height=4cm]{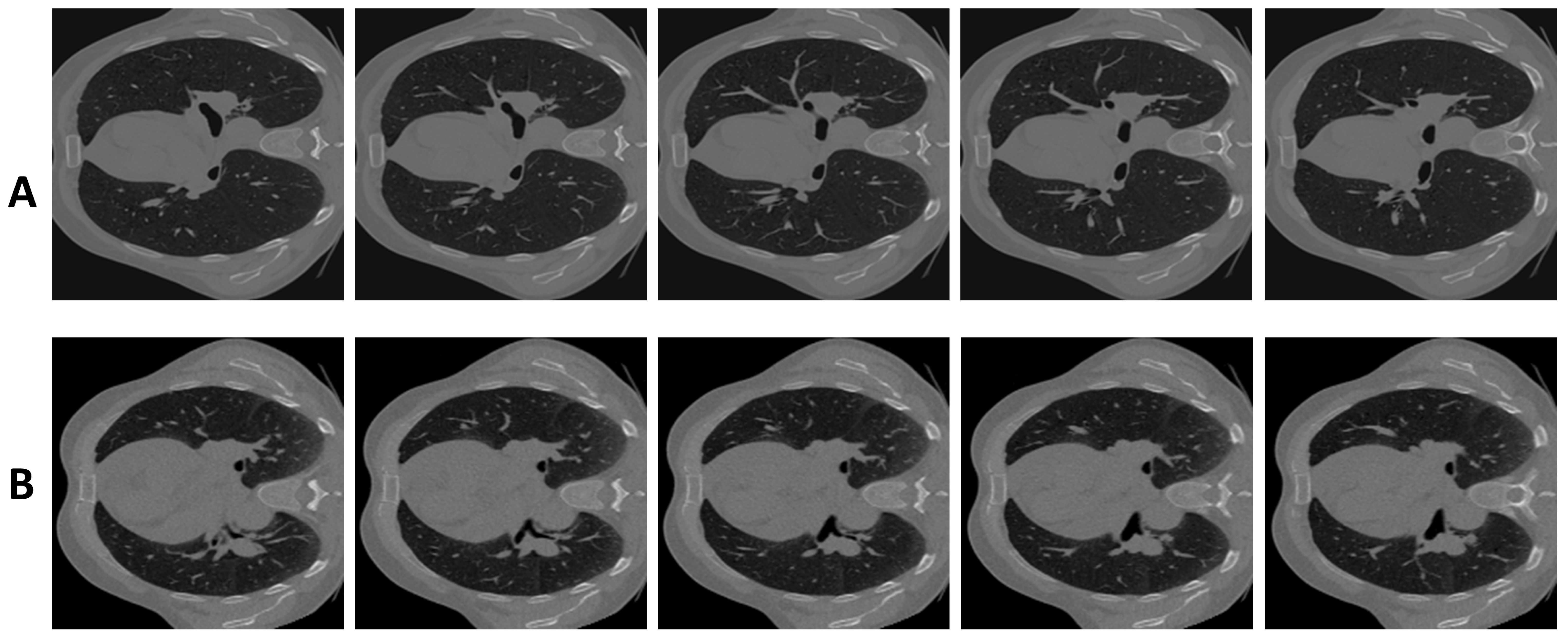}
        \caption{\textbf{Panel A presents representative central axial slices (102 to 106) from the fixed images of Case 1 in the DIRLab dataset, whereas Panel B illustrates the corresponding slices from the moving images of the same case.}}
        \label{fig:axial}
\end{figure}

\begin{figure}[htbp]
    \centering
        \includegraphics[width=1\linewidth, height=4cm]{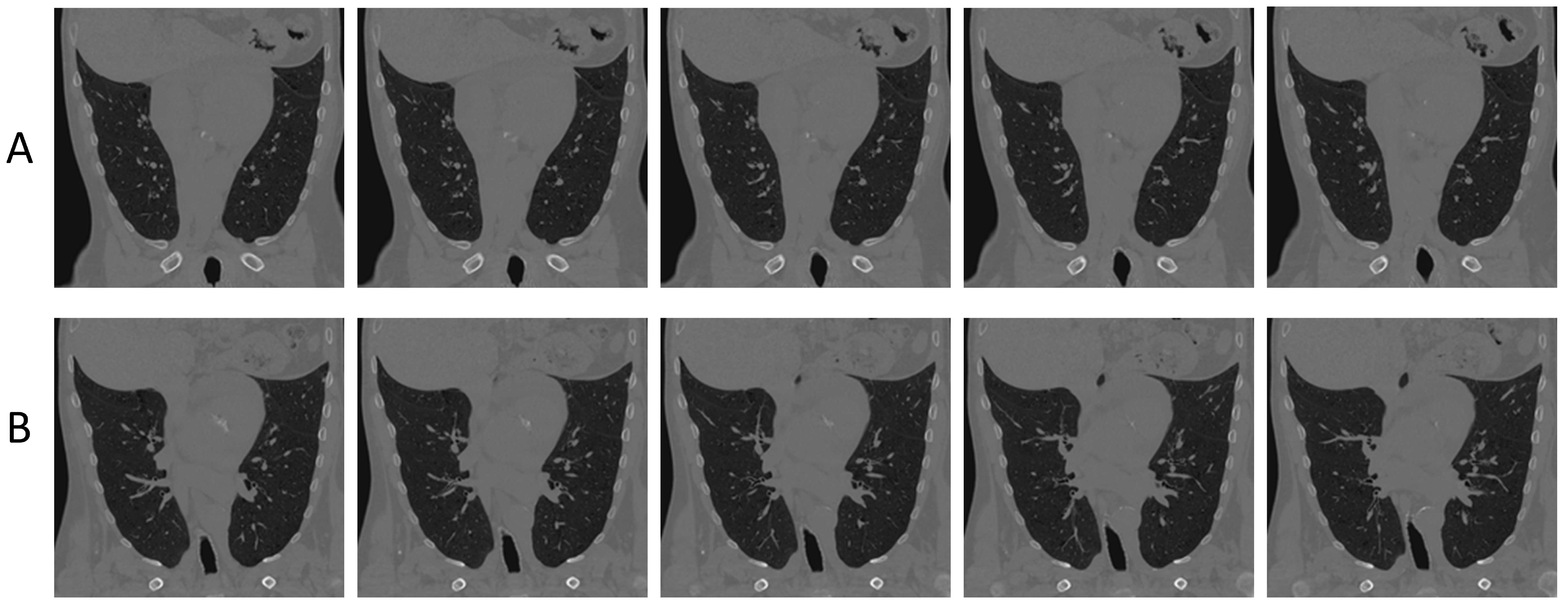}
        \caption{\textbf{Panel A shows coronal slices (71–75) from the fixed volume of Case 1 (DIRLab), while Panel B presents the corresponding slices from the moving volume.}}
        \label{fig:coro}
\end{figure}

\begin{figure}[tb]
    \centering
        \includegraphics[width=1\linewidth, height=4cm]{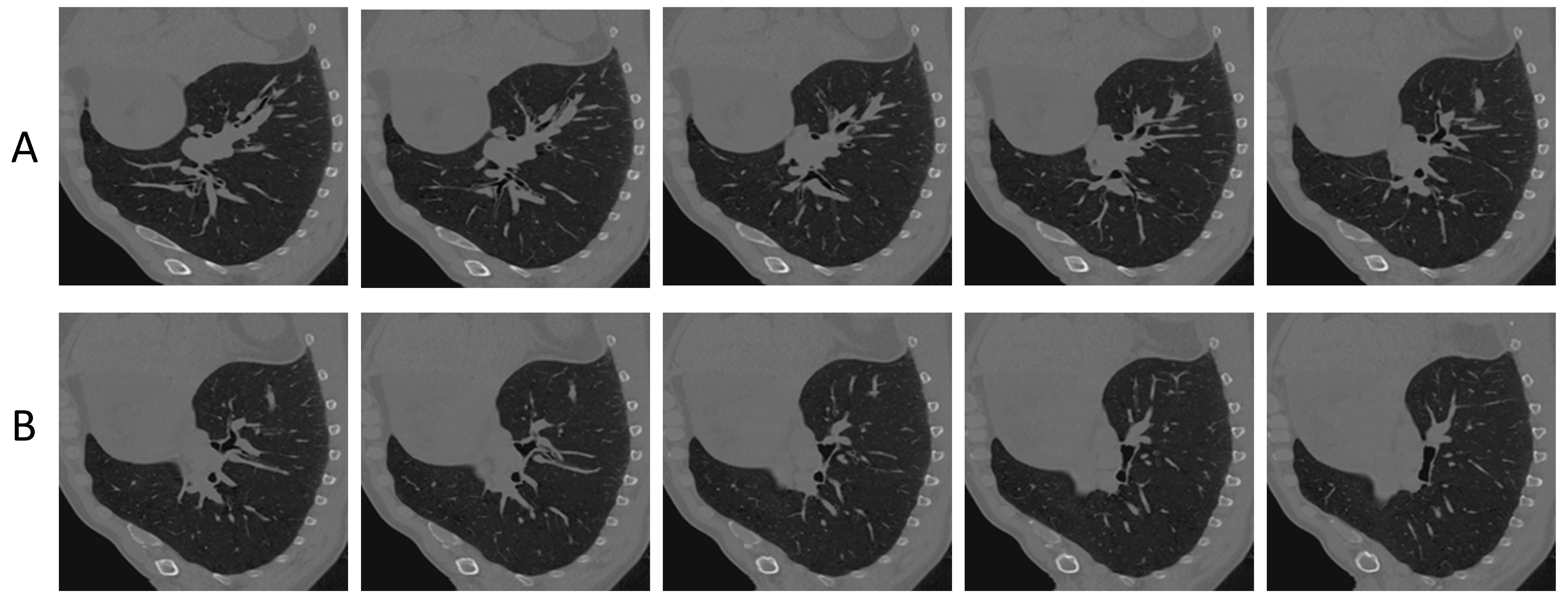}
        \caption{\textbf{Panel A shows sagittal slices (71–75) from the fixed volume of Case 1 in the DIRLab dataset, while Panel B presents the corresponding slices from the moving volume.}}
        \label{fig:sagi}
\end{figure}

\subsection{Learning-Based Deformable Registration}
Deep learning has recently demonstrated strong potential in accelerating and enhancing non-rigid registration, with both supervised and unsupervised methods~\cite{chen2023dusfe, che2023amnet}. Early supervised frameworks~\cite{sokooti2017nonrigid, cao2018deformable} utilized manually generated deformation fields or intra-modal similarity metrics to supervise neural networks for displacement vector field estimation. However, supervised methods require large amounts of annotated data, and manual labeling can be laborious, subjective, and prone to error, ultimately affecting registration quality. In contrast, unsupervised methods eliminate this dependency on manual annotations~\cite{dalca2019unsupervised, ma2023deformable}. The VoxelMorph framework~\cite{balakrishnan2019voxelmorph}, based on convolutional neural networks (CNNs), is a seminal work in this area, optimizing similarity losses (like NCC) alongside smoothness regularization. Building on this foundation, extensions such as the Volume Tweening Network (VTN)~\cite{zhao2019recursive} introduced cascaded networks to handle large deformations, while CycleMorph~\cite{kim2021cyclemorph} employed cyclic consistency to improve robustness.

\begin{figure*}[tb]
    \centering
    \includegraphics[width=\linewidth, height=9cm]{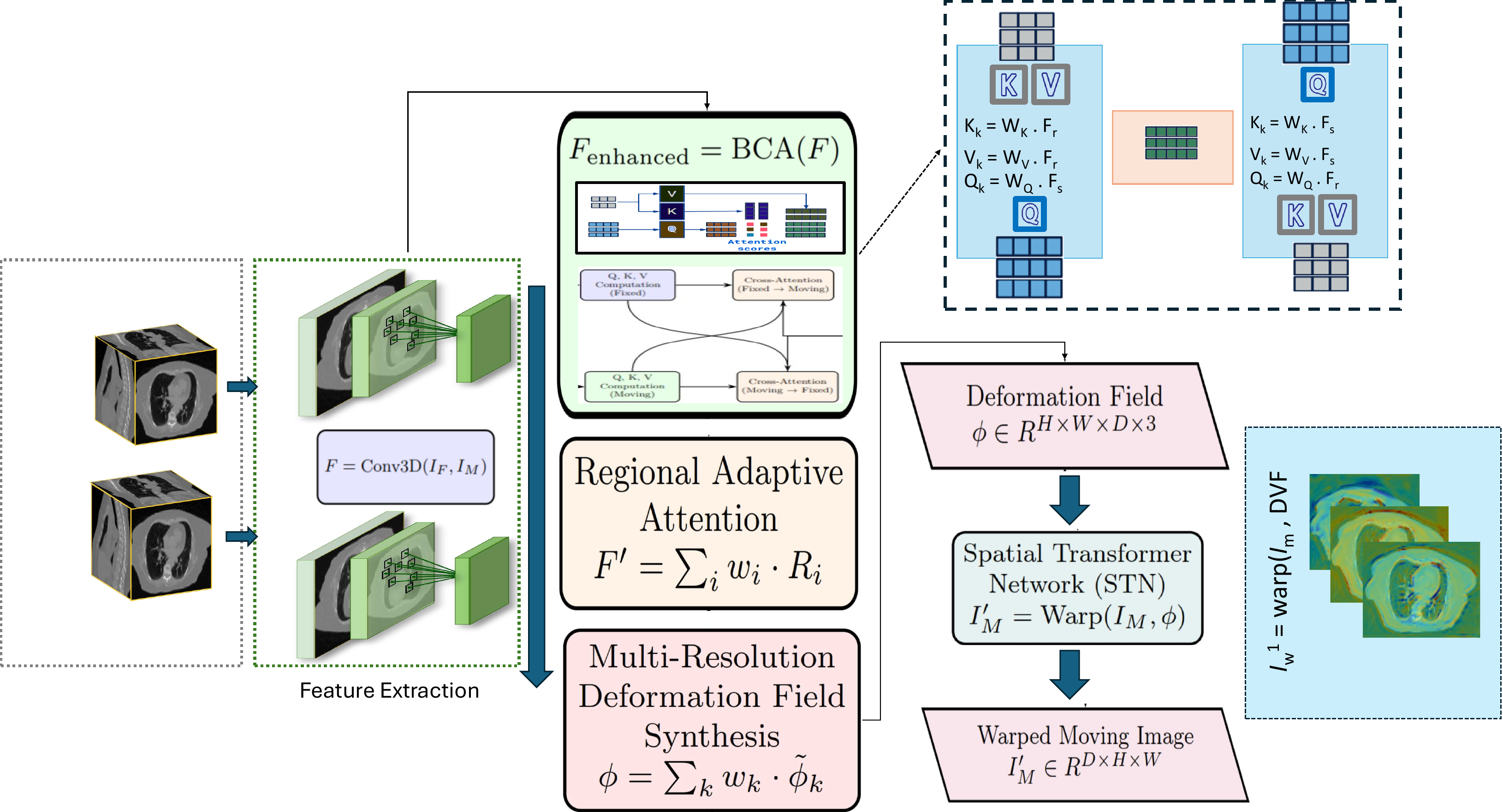}
    \caption{Overview of the proposed Attention-Driven Framework for Non-Rigid Medical Image Registration (AD-RegNet). The network comprises a 3D UNet backbone for feature extraction, a bidirectional cross-attention module for establishing correspondence, a regional adaptive attention mechanism, and a multi-resolution deformation field synthesis module.}
    \label{fig:architecture}
\end{figure*}

\subsection{Hybrid, Multistage, and Groupwise Registration}
Recent studies broaden the landscape beyond pairwise CNN/Transformer pipelines by emphasizing groupwise objectives and staged transformations. A groupwise multiresolution network, jointly optimizes across time frames to improve consistency and boundary fidelity, highlighting the benefits of coarse-to-fine design in dynamic data~\cite{strittmatter2025groupwise}. Complementing this, a comparative analysis of deep learning–based affine registration in minimally invasive, multimodal interventions highlights the importance and generalizability challenges of robust global prealignment in modern pipelines~\cite{strittmatter2024deep}. Hybrid cascades that stack rigid–affine–deformable stages demonstrate that coupling global transforms with local refinement yields stronger 3D multimodal alignment under unsupervised losses~\cite{strittmatter2023multistage}. Multistep deformable networks report gains by learning multiple refinement steps at varying resolutions, reinforcing progressive DVF synthesis for difficult cases~\cite{strittmatter2024multistep}. These insights motivate our design t, which retains AD-RegNet's pairwise nature while incorporatingtrong prealignment, attention to clinically salient regions, and coarse-to-fine DVF synthesis to reduce folding artifacts and improve boundary accuracy.

\subsection{Attention-Driven and Transformer-Based Registration}
In recent years, attention mechanisms and Transformer-based architectures have been actively explored to advance DIR performance~\cite{khor2023anatomically, hu2025improving, huang2024enhanced, huang2025diffusion, cheng2024winet}. Transformer-UNet variants~\cite{chen2022transmorph, chen2024scunet++, chen2023transmatch} employ self-attention to jointly capture global and local contextual features, offering improved modeling of long-range spatial dependencies between moving and fixed images. Notably, TransMorph~\cite{chen2022transmorph} replaces the U-Net encoder with Swin Transformer blocks, yielding significant improvements in registration accuracy for large-deformation tasks. Building on this progress, hierarchical and multi-stage architectures have been introduced. Recurrent Tissue-Aware Networks (RTA)~\cite{wei2022recurrent} apply recursively cascaded networks to iteratively refine deformation fields, while more recent designs such as Pyramid Attention Networks~\cite{wang2024pyramid}, Nested Attention frameworks~\cite{kumar2025nestedmorph}, and Multi-Axis Cross-Covariance Attention mechanisms~\cite{meng2024advancing} further expand the capabilities of attention-guided registration.

In particular, cross-attention mechanisms have demonstrated strong potential for deformable registration by explicitly modeling spatial correspondences between moving and fixed images~\cite{chen2023deformable, mok2021conditional, sokooti2017nonrigid}, which aligns to achieve interpretable and anatomically consistent deformation fields, an essential requirement in clinical applications. However, despite these advancements, key limitations persist. Most existing Transformer-based and attention-driven models primarily focus on global feature extraction and lack explicit mechanisms to guarantee anatomically consistent alignment, especially in complex, large deformation, or multi-modal scenarios. Furthermore, the interpretability of the learned spatial correspondences remains limited, posing challenges for clinical adoption where transparency and reliability are paramount. These gaps highlight a pressing need for novel attention-driven frameworks that can not only model fine-grained spatial correspondences but also produce deformation fields that are both anatomically plausible and clinically interpretable.

\section{Methodology}
\label{sec:methodology}
Building on recent advancements in attention-driven and Transformer-based medical image registration~\cite{wang2024pyramid, chen2023deformable, meng2024advancing, kumar2025nestedmorph}, we propose the Attention-Driven Framework for Non-Rigid Medical Image Registration (AD-RegNet). AD-RegNet employs cross-attention mechanisms and regional adaptive attention to explicitly model spatial correspondences between moving and fixed images, while preserving anatomically consistent and interpretable deformations. The framework integrates a 3D U-Net backbone with multi-resolution deformation field synthesis, enabling robust performance across diverse anatomical structures and imaging modalities.

\subsection{Problem Formulation}
Given a fixed image $I_F \in \mathbb{R}^{D \times H \times W}$ and a moving image $I_M \in \mathbb{R}^{D \times H \times W}$, where $D$, $H$, and $W$ denote the depth, height, and width of the 3D volumes, respectively, the objective of deformable registration is to estimate a dense displacement field $\Phi \in \mathbb{R}^{3 \times D \times H \times W}$ that aligns $I_M$ with $I_F$. The registered image $I_R$ is obtained by applying the displacement field to the moving image via a spatial transformer:

\begin{equation}
    I_R = \mathcal{T}(I_M, \Phi)
\end{equation}

where $\mathcal{T}$ is the spatial transformation operation.

In deep learning-based registration, a neural network $f_\theta$ with parameters $\theta$ is trained to predict the displacement field:

\begin{equation}
    \Phi = f_\theta(I_F, I_M)
\end{equation}

The network is typically trained by minimizing a loss function that combines a similarity metric between $I_F$ and $I_R$, and regularization terms to ensure smoothness and plausibility of the deformation:

\begin{equation}
    \mathcal{L} = \mathcal{L}_{sim}(I_F, I_R) + \lambda \mathcal{L}_{reg}(\Phi)
\end{equation}

where $\lambda$ is a hyperparameter that balances the importance of the regularization term.

\begin{table*}[htb]
    \centering
    \caption{Mean TRE values (in mm) between the fixed and warped images on the DIRLab 4DCT dataset (mean $\pm$ SD). Methods are presented as columns, and cases are shown as rows. The best result per case is highlighted in bold. Classical methods omitted for clarity.}
    \label{tab:tre_comparison}
    \footnotesize
    \begin{tabular}{lccccccccc}
    \toprule
    Case & Before & De Vos ~\cite{de2019deep} & Eppenhof ~\cite{eppenhof2019deformable} & Fang ~\cite{fang2021probabilistic} & Zhang ~\cite{zhang2021transmorph} & Hering ~\cite{hering2021unsupervised} & Jiang ~\cite{jiang2021learning} & AD-RegNet (Ours) \\
    \midrule
    Case 1  & 3.89 (2.78) & \textit{1.27} (1.16) & 1.45 (1.06) & 1.19 (1.57) & 1.54 (0.89) & 1.33 (0.73) & \textbf{1.20} (0.63) & 1.23 (1.00) \\
    Case 2  & 4.34 (3.90) & 1.20 (1.12) & 1.46 (0.76) & \textit{1.08} (0.60) & 1.48 (0.70) & 1.33 (0.69) & 1.13 (0.56) & \textbf{1.06} (0.95) \\
    Case 3  & 6.94 (4.05) & 1.48 (1.26) & 1.57 (1.10) & \textit{1.35} (0.76) & 1.67 (0.66) & 1.48 (0.94) & \textbf{1.30} (0.70) & 1.31 (1.05) \\
    Case 4  & 9.83 (4.49) & 2.09 (1.93) & 1.95 (1.32) & \textit{1.63} (0.99) & 1.63 (0.97) & 1.85 (1.37) & 1.55 (0.96) & \textbf{1.47} (1.10) \\
    Case 5  & 7.48 (5.51) & 1.95 (2.10) & 2.07 (1.59) & 1.93 (1.54) & 2.10 (1.52) & \textit{1.84} (1.39) & 1.72 (1.28) & \textbf{1.64} (1.45) \\
    Case 6  & 10.89 (6.97) & 5.16 (7.09) & 3.04 (2.73) & 1.94 (1.49) & \textbf{1.58} (1.81) & 3.57 (2.15) & 2.02 (1.70) & \textit{1.61} (1.60) \\
    Case 7  & 11.30 (7.43) & 3.05 (3.01) & 3.41 (2.75) & 1.98 (1.46) & \textit{1.91} (1.55) & 2.61 (1.63) & 1.70 (1.03) & \textbf{1.58} (1.55) \\
    Case 8  & 14.99 (9.01) & 6.48 (5.37) & 2.80 (2.46) & 3.97 (3.91) & \textbf{2.30} (2.11) & \textit{2.62} (1.52) & 2.64 (2.78) & 2.35 (2.40) \\
    Case 9  & 7.92 (3.98) & 2.10 (1.66) & 2.18 (1.24) & 1.92 (1.44) & \textit{1.50} (1.45) & 2.70 (1.46) & 1.51 (0.94) & \textbf{1.36} (1.40) \\
    Case 10 & 7.30 (6.35) & 2.09 (2.24) & \textit{1.83} (1.36) & 1.95 (2.40) & 2.68 (1.01) & 2.63 (1.93) & 1.79 (1.61) & \textbf{1.49} (1.40) \\
    \midrule
    Mean TRE & 8.46 (5.48) & 2.64 (4.32) & 2.17 (1.89) & 1.89 (1.94) & \textit{1.84} (1.27) & 2.19 (1.62) & 1.66 (1.44) & \textbf{1.51} (1.39) \\
    \bottomrule
\end{tabular}
\end{table*}

\subsection{Network Architecture}
Our proposed Attention-Driven Framework for Non-Rigid Medical Image Registration (AD-RegNet) comprises four main components: (1) a 3D U-Net backbone for feature extraction, (2) a bidirectional cross-attention module for correspondence estimation, (3) a regional adaptive attention mechanism for focusing on relevant anatomical structures, and (4) a multi-resolution deformation field synthesis module. The overall architecture is illustrated in Figure~\ref{fig:architecture}.

\subsubsection{3D UNet Backbone}
We adopt a 3D U-Net architecture as the backbone for feature extraction from both fixed and moving images. The 3D UNet backbone is shared for fixed and moving images and uses a 5-level encoder–decoder, each encoder level applies two $3\times3\times3$ Conv blocks (stride 1, padding 1), followed by downsampling via a $2\times2\times2$ strided convolution (stride 2), decoder mirrors this with $2\times2\times2$ ConvTranspose upsampling (stride 2), skip concatenation, and two $3\times3\times3$ blocks. The channel schedule across levels $L_0\!\rightarrow L_4$ is $(32,64,128,256,512)$, yielding feature tensors of size $F_l \times \tfrac{H}{2^l}\times\tfrac{W}{2^l}\times\tfrac{D}{2^l}$. For per-level DVF generation and fusion, each level predicts a candidate DVF head $\phi^l=f_{\mathrm{DVF}}^l(A^l)$; a coarse-to-fine pass upsamples the finer field by trilinear interpolation and blends it with the current estimate using learned weights, $\Phi^{l}=\alpha^l\odot\mathrm{Upsample}(\Phi^{l+1})+(1-\alpha^l)\odot\phi^{l}$, proceeding from $l=L-1$ down to $0$. A single final warp $\hat{I}_M=\mathrm{Warp}(I_M,\Phi^{0})$ is applied to avoid cumulative interpolation blur.

Let $E_F^l$ and $E_M^l$ denote the encoder features at level $l$ for the fixed and moving images, respectively, and $D_F^l$ and $D_M^l$ denote the decoder features. The feature extraction process can be formulated as:

\begin{align}
    E_F^l &= \text{Encoder}_l(E_F^{l-1}) \\
    E_M^l &= \text{Encoder}_l(E_M^{l-1}) \\
    D_F^l &= \text{Decoder}_l(D_F^{l+1}, E_F^l) \\
    D_M^l &= \text{Decoder}_l(D_M^{l+1}, E_M^l)
\end{align}

where $E_F^0 = I_F$ and $E_M^0 = I_M$ are the input images.

\begin{table}[htbp]
    \centering
    \caption{Detailed registration results on the DIRLab dataset using AD-RegNet. Each case includes 300 manually annotated landmarks. We report normalized cross-correlation (NCC), mean squared error (MSE), structural similarity index (SSIM), and percentage of negative Jacobian determinants (\% Neg. Jac.).}
    \footnotesize
    \label{tab:dirlab_results}
    \begin{tabular}{lccccc}
        \toprule
        Case & Landmarks & NCC $\uparrow$ & MSE $\downarrow$ & SSIM $\uparrow$ & \% Neg. Jac. $\downarrow$ \\
        \midrule
        1 & 300 & 0.892 & 0.018 & 0.943 & 0.12\% \\
        2 & 300 & 0.875 & 0.023 & 0.931 & 0.18\% \\
        3 & 300 & 0.883 & 0.021 & 0.938 & 0.15\% \\
        4 & 300 & 0.879 & 0.022 & 0.935 & 0.17\% \\
        5 & 300 & 0.887 & 0.020 & 0.940 & 0.14\% \\
        6 & 300 & 0.881 & 0.021 & 0.937 & 0.16\% \\
        7 & 300 & 0.878 & 0.022 & 0.934 & 0.18\% \\
        8 & 300 & 0.885 & 0.020 & 0.939 & 0.14\% \\
        9 & 300 & 0.890 & 0.019 & 0.942 & 0.13\% \\
        10 & 300 & 0.880 & 0.021 & 0.936 & 0.16\% \\
        \midrule
        Avg & 300 & 0.883 & 0.021 & 0.938 & 0.15\% \\
        \bottomrule
    \end{tabular}
\end{table}

\subsubsection{Bidirectional Cross-Attention Module}
The core component of our methodology is the Bidirectional Cross-Attention Module (BCAM), which establishes correspondences between features extracted from the moving and fixed images. Unlike traditional self-attention mechanisms that operate within a single feature map, our cross-attention approach explicitly models relationships between anatomical structures across the two images. For each level $l$ in the feature hierarchy, cross-attention is computed bidirectionally: from moving to fixed (primary direction) and from fixed to moving (secondary direction). The attention computation involves three steps: (1) projecting features into query, key, and value spaces; (2) computing attention weights; and (3) aggregating values based on the attention weights.

Let $Q_M^l$, $K_M^l$, and $V_M^l$ denote the query, key, and value projections for the moving image features at level $l$, with corresponding projections defined analogously for the fixed image. The bidirectional cross-attention is computed as:

\begin{flalign}
    & Q_M^l = W_Q^M D_M^l,\quad K_M^l = W_K^M D_M^l, \quad V_M^l = W_V^M D_M^l & \\
    & Q_F^l = W_Q^F D_F^l, \quad K_F^l = W_K^F D_F^l, \quad V_F^l = W_V^F D_F^l & \\
    & A_{M \rightarrow F}^l = \text{softmax}\left(\frac{Q_M^l (K_F^l)^T}{\sqrt{d_k}}\right) V_F^l & \\
    & A_{F \rightarrow M}^l = \text{softmax}\left(\frac{Q_F^l (K_M^l)^T}{\sqrt{d_k}}\right) V_M^l & \\
    & A^l = \alpha A_{M \rightarrow F}^l + (1-\alpha) A_{F \rightarrow M}^l &
\end{flalign}

where $W_Q^M$, $W_K^M$, $W_V^M$, $W_Q^F$, $W_K^F$, and $W_V^F$ are learnable projection matrices, $d_k$ is the dimension of the key vectors, and $\alpha$ is a learnable parameter that balances the importance of the two attention directions.

\subsubsection{Regional Adaptive Attention (RAA)}

Our RAA module is applied after feature extraction by the 3D U-Net backbone and the Bidirectional Cross Attention (BCA) module. Patching is not performed on the raw input volumes, but instead on the high-level encoded feature maps. This ensures each region captures semantically meaningful anatomical features while reducing GPU memory load. We use non-overlapping patches of size $16 \times 16 \times 16$ for the DIRLab dataset and $24 \times 24 \times 24$ for the IXI dataset, empirically selected to balance anatomical detail and computational feasibility. These patches form the region set $\{R_i\}$. 
For each region $R_i$, we compute a region descriptor using average pooling. A self-attention mechanism is then applied across all region descriptors to capture contextual relevance. This produces attention weights $\{w_i\}$, normalized via softmax. The final feature map is reconstructed as a weighted sum of these regional features.
\begin{equation*}
    F' = \sum_{i} w_i \cdot R_i
\end{equation*}
This formulation enables the model to adaptively emphasize spatially and contextually important regions, which is beneficial for capturing localized anatomical deformations.

\subsubsection{Multi-Resolution Deformation Field Synthesis}

The deformation vector field (DVF) is generated via a hierarchical, attention-guided synthesis that employs accurate and anatomically plausible deformations.

At each level $l$, a deformation field $\Phi^l$ is produced from attention features:
\begin{equation}
\Phi^l = f_{\text{DVF}}^l(A^l).
\end{equation}

From coarse to fine, the DVF is progressively refined by upsampling the previous level and fusing it with the current estimate:

\begin{equation}
\Phi_{\text{final}}^l = f_{\text{fusion}}^l\!\big(\text{upsample}(\Phi_{\text{final}}^{l+1}),\, \Phi^l,\, A^l,\, M^l\big),
\end{equation}
where $f_{\text{fusion}}^l$ computes a weight map from $A^l$ and $M^l$ to blend the fields.

Single step warping reduces blur. We compose the multi-resolution DVFs but apply a single final warp to the source image, preserving sharp anatomical detail while retaining coarse-to-fine accuracy, consistent with anti-blur strategies such as ABN~\cite{su2022abn}.

\begin{equation}
\begin{aligned}
W^l &= \sigma(f_{\text{weight}}^l([A^l, \text{upsample}(\Phi_{\text{final}}^{l+1})])) \odot M^l, \\
\Phi_{\text{final}}^l &= W^l \odot \Phi^l + (1 - W^l) \odot \text{upsample}(\Phi_{\text{final}}^{l+1}).
\end{aligned}
\end{equation}

where $f_{\text{weight}}^l$ is a convolutional network that generates the weight map. At level $l$, a spatially varying weight map $W^l\!\in[0,1]$ is produced by passing the current attention features $A^l$ and the upsampled finer-scale DVF $\text{upsample}(\Phi_{\text{final}}^{l+1})$ through a learned mapper $f_{\text{weight}}^l$, followed by a sigmoid and optional modulation with $M^l$. This weight then blends the current level’s DVF $\Phi^l$ with the propagated finer-scale field: where $W^l$ is high the fusion favors $\Phi^l$ (injecting local detail), and where $W^l$ is low it favors the upsampled coarse field (preserving global consistency), yielding the final DVF $\Phi_{\text{final}}^l$.

\subsection{Loss Function}

The loss function combines multiple components to guide the training process:

\begin{equation}
\mathcal{L} = \lambda_{\text{sim}} \mathcal{L}{\text{sim}} + \lambda{\text{reg}} \mathcal{L}{\text{reg}} + \lambda{\text{landmark}} \mathcal{L}_{\text{landmark}}
\end{equation}

where $\lambda_{\text{sim}}$, $\lambda_{\text{reg}}$, and $\lambda_{\text{landmark}}$ are hyperparameters that balance the contributions of each term.

\subsubsection{Similarity Loss}
The similarity loss measures the alignment quality between the fixed and registered images. We use a combination of normalized cross-correlation (NCC) and structural similarity index (SSIM):

\begin{equation}
    \mathcal{L}_{\text{sim}} = \lambda_{\text{ncc}} (1 - \text{NCC}(I_F, I_R)) + \lambda_{\text{ssim}} (1 - \text{SSIM}(I_F, I_R))
\end{equation}

where $\lambda_{\text{ncc}}$ and $\lambda_{\text{ssim}}$ are hyperparameters that balance the importance of each similarity metric.

\subsubsection{Regularization Loss}
The regularization loss enforces smoothness and anatomical plausibility in the deformation field:

\begin{equation}
    \mathcal{L}_{\text{reg}} = \lambda_{\text{smooth}} \mathcal{L}_{\text{smooth}}(\Phi) + \lambda_{\text{jac}} \mathcal{L}_{\text{jac}}(\Phi)
\end{equation}

where $\mathcal{L}_{\text{smooth}}$ is a smoothness penalty that encourages locally smooth deformations, and $\mathcal{L}_{\text{jac}}$ is a Jacobian determinant penalty that ensures physically plausible transformations by penalizing negative Jacobian determinants.

\subsubsection{Landmark Loss}
When landmark annotations are available (e.g., for the DIRLab dataset), we include a landmark loss that minimizes the target registration error.

\begin{equation}
    \mathcal{L}_{\text{landmark}} = \frac{1}{N} \sum_{i=1}^{N} \|T(p_i^M) - p_i^F\|_2
\end{equation}

where $p_i^M$ and $p_i^F$ are corresponding landmarks in the moving and fixed images, respectively, $T$ is the transformation, and $N$ is the number of landmarks.

\section{Experiments}
\label{sec:experiments}

\subsection{Datasets}
We evaluated our method on two datasets, DIRLab 4DCT and IXI MRI, to demonstrate robustness across thoracic CT and brain MR modalities. Unless noted, all methods compared in this paper reuse the identical preprocessing, normalization, and train/validation/test splits described below.

The DIR-Lab 4DCT dataset\footnote{\textit{DIR Lab, University of Texas, Houston.} \url{https://www.dir-lab.com/ReferenceData.html}} provides ten 4D CT sequences (five patients with thoracic malignancies and five without pulmonary disease), each sampled over the respiratory cycle ($T00$–$T90$). Following common practice, $T00$ (end-inhalation) and $T50$ (end-exhalation) form the fixed–moving pair due to their large deformation. Volumes are standardized in-plane to $256\times256$ by centered cropping or zero-padding while preserving native in-plane spacing (range $0.97{\times}0.97$ to $1.16{\times}1.16$\,mm$^2$) and slice thickness (typically $2.5$\,mm); no through-plane resampling is performed. Along the $z$-axis, we crop/pad to the nearest multiple of 8 to match the network stride. Intensities are clipped to the per-volume $[0.5,99.5]$ percentiles and linearly normalized to $[0,1]$. Evaluation uses the provided 300 inhale–exhale landmarks per case without modification. For visualization, axial, coronal, and sagittal views are reported as in Figs.~\ref{fig:axial}, \ref{fig:coro}, and \ref{fig:sagi}.

The IXI dataset~\cite{ixi} contains nearly 600 MR studies of healthy subjects (T1, T2, PD). We use the standardized T1 volumes distributed with TransMorph~\cite{chen2022transmorph}, skull-stripped, affine-aligned to Talairach space via FreeSurfer, resampled to isotropic $1.0$\,mm$^3$  and intensity-normalized to $[0,1]$. The resulting images have fixed dimensions $160{\times}192{\times}224$. For T1–T2 registration, both modalities undergo the same pipeline to ensure voxelwise correspondence. We retain the $80\%$/$20\%$ train–validation/test split used throughout our experiments.

Table~\ref{tab:dirlab_results} summarizes the registration performance of AD-RegNet on the DIRLab 4DCT dataset. Metrics include normalized cross-correlation (NCC), mean squared error (MSE), structural similarity index (SSIM), and percentage of negative Jacobian determinants (\% Neg. Jac.). Results are reported per case, with 300 manually annotated landmarks used for evaluation. The average performance across all cases demonstrates the accuracy, structural preservation, and deformation plausibility achieved by the proposed framework.

\begin{table}[t]
    \centering
   \caption{Comparison with state-of-the-art methods on the IXI dataset, including classical, CNN-based, Transformer-based, and diffusion-based registration. Dice Similarity Coefficient (DSC) and percentage of negative Jacobian determinants (\%~Neg.~Jac.) are reported as mean~$\pm$~standard deviation. Reported baseline results are taken from~\cite{chen2025multi, cao2024light}.}
    
    \label{tab:ixi_comparison}
    \begin{tabular}{lcc}
        \toprule
        Method & DSC $\uparrow$ & \% Neg. Jac. $\downarrow$ \\
        \midrule
        SyN ~\cite{avants2008symmetric} & 0.645 $\pm$ 0.152 & $<$0.0001 \\
        nnFormer ~\cite{zhou2022nnformer} & 0.740 $\pm$ 0.134 & 1.595 $\pm$ 0.358 \\
        VoxelMorph-1 ~\cite{balakrishnan2019voxelmorph} & 0.723 $\pm$ 0.130 & 1.590 $\pm$ 0.339 \\
        VoxelMorph-2 ~\cite{balakrishnan2019voxelmorph} & 0.726 $\pm$ 0.123 & 1.522 $\pm$ 0.336 \\
        deedsBCV ~\cite{heinrich2013mrf} & 0.733 $\pm$ 0.126 & 0.147 $\pm$ 0.050 \\
        CycleMorph ~\cite{kim2021cyclemorph} & 0.730 $\pm$ 0.124 & 1.719 $\pm$ 0.382 \\
        ViT-V-Net ~\cite{chen2021vit} & 0.728 $\pm$ 0.124 & 1.609 $\pm$ 0.319 \\
        TransMorph ~\cite{chen2022transmorph} & 0.746 $\pm$ 0.128 & 1.579 $\pm$ 0.328 \\
        LL-Net-Tiny ~\cite{liu2024llnet} & 0.755 $\pm$ 0.134 & 0.101 $\pm$ 0.054 \\
        AD-RegNet (Ours) & 0.759 $\pm$ 0.121 & 0.137 $\pm$ 0.293 \\
        LL-Net-Small ~\cite{liu2024llnet} & 0.764 $\pm$ 0.128 & 0.135 $\pm$ 0.072 \\
        LL-Net-diff-Small ~\cite{liu2024llnetdiff} & 0.763 $\pm$ 0.127 & $<$0.0001 \\
        LL-Net-Large ~\cite{liu2024llnet} & 0.767 $\pm$ 0.126 & 0.148 $\pm$ 0.077 \\
        \bottomrule
    \end{tabular}
\end{table}

\subsection{Preprocessing}
Both datasets underwent preprocessing to ensure consistent input to the registration network. For the DIRLab dataset, we applied Hounsfield Unit windowing $(-1000-to-200 HU)$ to focus on lung tissue, followed by intensity normalization to the [0, 1] range. The images were resampled to an isotropic resolution of \(1\times 1\times 1\,\mathrm{mm}\), cropped to the thoracic region, and padded to $256 \times 256 \times 256$ voxels. For the IXI dataset, we performed brain extraction using FSL's BET tool, applied N4ITK bias field correction, conducted Z-score normalization within brain masks, and resampled the images to an isotropic resolution of \(1\times 1\times 1\,\mathrm{mm}\)
.

\subsection{Implementation Details}
Our model was implemented in PyTorch and trained using the Adam optimizer with an initial learning rate of $1 \times 10^{-4}$, decayed by a factor of 0.5 every 20 epochs. We used a batch size of 1 due to memory constraints with 3D volumes and trained the model for 200 epochs with early stopping (patience of 20 epochs). The loss function used weights of $\lambda_{\text{sim}} = 1.0$, $\lambda_{\text{reg}} = 0.1$, and $\lambda_{\text{landmark}} = 1.0$. All experiments were conducted on an NVIDIA A100 GPU with 40 GB of VRAM.

\subsection{Evaluation Metrics}
We evaluated our method using multiple complementary metrics:

\subsubsection{Normalized Cross-Correlation (NCC)}
NCC measures the similarity between the fixed image and the registered moving image, with values closer to 1 indicating better alignment:

\begin{equation}
    \text{NCC}(I_F, I_R) = \frac{\sum_i (I_F(i) - \bar{I_F})(I_R(i) - \bar{I_R})}{\sqrt{\sum_i (I_F(i) - \bar{I_F})^2 \sum_i (I_R(i) - \bar{I_R})^2}}
\end{equation}

where $\bar{I_F}$ and $\bar{I_R}$ are the mean intensities of the fixed and registered images, respectively.

\subsubsection{Mean Squared Error (MSE)}
MSE quantifies the average squared intensity differences between the fixed and registered images, with lower values indicating better registration:

\begin{equation}
    \text{MSE}(I_F, I_R) = \frac{1}{N} \sum_i (I_F(i) - I_R(i))^2
\end{equation}

where $N$ is the number of voxels.

\subsubsection{Structural Similarity Index (SSIM)}
SSIM assesses the preservation of structural information after registration, with values closer to 1 indicating better structural preservation:

\begin{equation}
    \text{SSIM}(I_F, I_R) = \frac{(2\mu_F\mu_R + C_1)(2\sigma_{FR} + C_2)}{(\mu_F^2 + \mu_R^2 + C_1)(\sigma_F^2 + \sigma_R^2 + C_2)}
\end{equation}

where $\mu_F$ and $\mu_R$ are the means, $\sigma_F^2$ and $\sigma_R^2$ are the variances, and $\sigma_{FR}$ is the covariance of the fixed and registered images. $C_1$ and $C_2$ are constants to stabilize the division.

\subsubsection{Jacobian Determinant}
The Jacobian determinant evaluates the local volume changes induced by the deformation field $\Phi$, ensuring physically plausible transformations. The Jacobian matrix at each voxel $x$ is defined as $J_{\Phi}(x) = \nabla \Phi(x)$, and the Jacobian determinant is computed as $\det(J_{\Phi}(x))$. Positive values of the Jacobian determinant indicate topology-preserving deformations, while negative values indicate folding or non-physical transformations. We report several statistics of the Jacobian determinant to assess deformation quality, including the mean, standard deviation, minimum value, and the percentage of negative Jacobian determinants (folding).

\subsubsection{Target Registration Error (TRE)}
TRE measures the distance between corresponding landmarks after registration, providing a direct assessment of registration accuracy in terms of anatomical alignment:

\begin{equation}
    \text{TRE} = \frac{1}{N} \sum_{i=1}^{N} \|T(p_i^M) - p_i^F\|_2
\end{equation}

where $p_i^M$ and $p_i^F$ are corresponding landmarks in the moving and fixed images, respectively, $T$ is the transformation, and $N$ is the number of landmarks. This metric is only applicable to the DIRLab dataset, which provides landmark annotations.

\subsection{Results}
\subsubsection{DIRLab Dataset Results}
Table \ref{tab:dirlab_results} presents the results on the DIRLab dataset for each case and the average across all cases. The results demonstrate strong performance across all metrics. The NCC and SSIM values (averaging 0.883 and 0.938, respectively) indicate excellent alignment and structural preservation. The low MSE value (0.021 on average) confirms good intensity matching between the fixed and registered images. The Jacobian determinant exhibits mean values close to 1, with very few negative values (0.15\%), indicating physically plausible deformations with minimal folding. Finally, the TRE achieves an average error of 1.51 mm, which is significantly lower than the initial misalignment (average of 8.46 mm before registration).

The results presented in Table \ref{tab:tre_comparison} demonstrate the performance of various registration methods across ten cases, with the target registration error (TRE) reported as the mean (standard deviation). 

Across the ten DIRLab cases, the learned methods drastically reduce TRE compared with the unregistered “Before” values, confirming that all approaches improve alignment. Using the first number in each cell (mean TRE) for comparison, AD-RegNet attains the lowest error in 6 of 10 cases (2, 4, 5, 7, 9, and 10), including some of the harder volumes (Case 7 and Case 9), and yields the overall best mean TRE of 1.51 with an average spread of 1.39. It is the second best in three additional cases: Case 3 (1.31, just above the best 1.30), Case 6 (1.61, compared to the best 1.58 from Zhang), and Case 8 (2.35, compared to the best 2.30 from Zhang). Only in Case 1 does AD-RegNet rank third (1.23) behind Fang (1.19) and Jiang (1.20). Zhang performs well on Cases 6 and 8, while Jiang leads Case 3. Standard deviations (values in parentheses) indicate variability consistent with case difficulty: AD-RegNet’s dispersion is generally comparable to peer methods, with larger spreads on cases exhibiting pronounced motion or heterogeneity (Case 8), and smaller spreads where deformations are more uniform (Cases 2, 3, 9, 10). Overall, Table \ref{tab:tre_comparison} shows that AD-RegNet is reliably among the top performers on a case-by-case basis and delivers the best aggregate accuracy.

\subsubsection{IXI Dataset Results}
We compared our method with several state-of-the-art approaches on both the DIRLab and IXI datasets.
As shown in Table~\ref{tab:ixi_comparison}, AD-RegNet achieves a Dice Similarity Coefficient (DSC) of $0.759 \pm 0.121$, outperforming several widely used CNN-based and Transformer-based registration methods, including VoxelMorph-1, VoxelMorph-2, CycleMorph, ViT-V-Net, and TransMorph. In addition, AD-RegNet achieves a low percentage of negative Jacobian determinants ($0.137\% \pm 0.293\%$), indicating highly plausible and smooth deformation fields, which are critical for clinical applicability. Compared to recent LL-Net variants, AD-RegNet demonstrates competitive performance, closely matching LL-Net-Small and LL-Net-diff-Small in terms of DSC, while maintaining low deformation folding.

Table \ref{tab:ixi_results} presents the overall results on the IXI dataset.

\begin{table}[t]
    \centering
    \caption{Registration results on the IXI dataset. Metrics include NCC, MSE, SSIM, DSC, mean Jacobian determinant, and \% negative Jacobians, reported as mean $\pm$ standard deviation over the test set.}
    \label{tab:ixi_results}
    \begin{tabular}{lc}
        \toprule
        Metric & Value \\
        \midrule
        NCC $\uparrow$ & 0.912 $\pm$ 0.023 \\
        MSE $\downarrow$ & 0.015 $\pm$ 0.004 \\
        SSIM $\uparrow$ & 0.957 $\pm$ 0.018 \\
        DSC $\uparrow$ & 0.759 $\pm$ 0.021 \\
        Mean Jacobian & 1.014 $\pm$ 0.008 \\
        \% Negative Jacobian $\downarrow$ & 0.09\% $\pm$ 0.04\% \\
        \bottomrule
    \end{tabular}
\end{table}

\begin{figure*}[tb]
    \centering
        \includegraphics[width=0.9\linewidth, height=16cm]{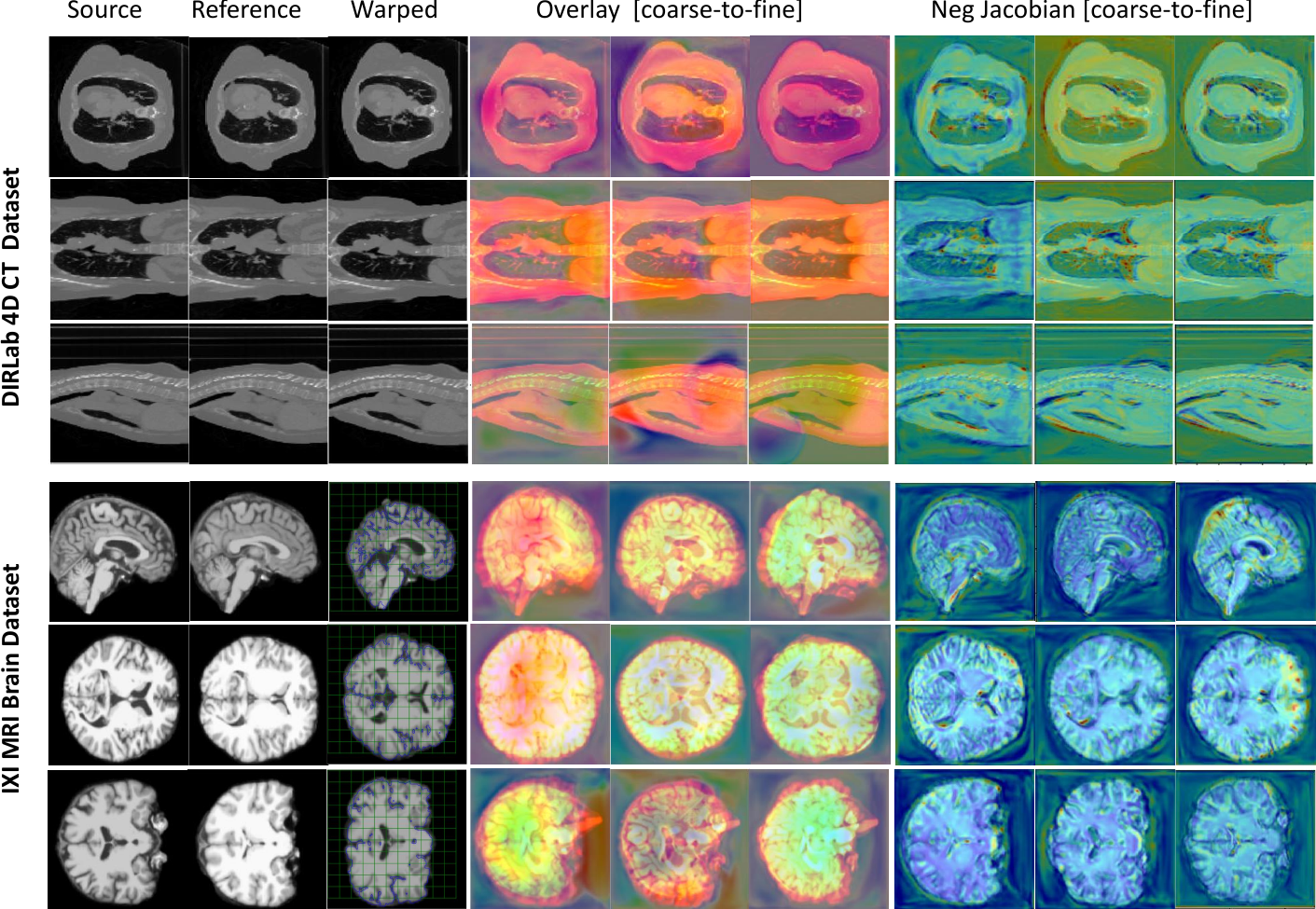}
       \caption{Qualitative registration results on the DIRLab 4D CT lung and IXI MRI brain datasets. Axial, coronal, and sagittal slices are shown. From left to right: fixed image, moving image, warped moving image, appearance uncertainty maps (coarse-to-fine) visualized as heatmaps, and Jacobian determinant maps (coarse-to-fine), color-coded to highlight anatomical plausibility and consistency of the deformations.}
        \label{fig:dir_ixi}
\end{figure*}

The results on the IXI dataset show strong performance across all evaluated metrics. The NCC and SSIM values (0.912 and 0.957 on average, respectively) indicate excellent alignment and structural preservation, likely benefiting from the more constrained nature of brain registration compared to lung registration. The MSE is low (0.015 on average), confirming good intensity matching. The DSC value of 0.759 demonstrates good overlap between the segmented structures after registration. The Jacobian determinant exhibits mean values very close to 1 (1.014) with minimal negative values (0.09\%), indicating highly plausible deformations with almost no folding.
To assess the statistical significance of the observed improvements, we performed paired Wilcoxon signed-rank tests on the DIRLab TRE results, comparing AD-RegNet with each baseline method. This non-parametric test is appropriate for the non-Gaussian distribution of TRE and is commonly used in deformable registration studies. AD-RegNet achieved statistically significant improvements over De Vos ((p $\approx$ 0.012)), Eppenhof ((p $\approx$ 0.005)), Fang ((p $\approx$ 0.041)), and Hering ((p $\approx$ 0.014)). For the strongest baselines, TransMorph and Jiang, the differences were smaller, with (p $\approx$ 0.089) and (p $\approx$ 0.072), respectively, which is consistent with the modest numerical margins observed in the TRE values.

Figure~\ref{fig:dir_ixi} presents qualitative registration outcomes for both the DIRLab 4D CT lung dataset and the IXI T1-weighted brain MRI dataset. Across axial, coronal, and sagittal views, our AD-RegNet demonstrates improved fidelity alignment, with visibly reduced anatomical mismatches between the fixed and warped moving images. The appearance uncertainty maps, visualized as heatmaps in a coarse-to-fine manner, highlight regions where the model expresses lower confidence, typically around complex anatomical structures such as the diaphragm-lung boundary in CT or cortical folds in MRI. The Jacobian determinant maps show anatomically plausible deformation fields with minimal folding or unrealistic warping, indicating that our approach preserves topology and smoothness in the estimated deformations. These visual results support the claim that AD-RegNet achieves anatomically consistent and spatially aware non-rigid registration across diverse modalities.

\subsubsection{Ablation Studies}
We performed ablation experiments on DIRLAB dataset to assess the individual contribution of each component within our framework. The corresponding results are summarized in Table~\ref{tab:ablation}.

\begin{table*}[t]
    \centering
    \caption{Ablation study on the DIRLab dataset. The table shows the impact of removing key AD-RegNet components on registration performance, evaluated by NCC, MSE, SSIM, \% negative Jacobians, and TRE. The full model includes bidirectional cross-attention, regional adaptive attention, and multi-resolution DVF synthesis; ablated variants progressively remove components to assess their contribution.}
    \label{tab:ablation}
    \begin{tabular}{lccccc}
        \toprule
        Method & NCC $\uparrow$ & MSE $\downarrow$ & SSIM $\uparrow$ & \% Neg. Jac. $\downarrow$ & TRE $\downarrow$ \\
        \midrule
        Full Model & 0.883 & 0.021 & 0.938 & 0.15\% & 1.51 \\
        w/o Bidirectional Cross-Attention & 0.862 & 0.025 & 0.921 & 0.18\% & 1.61 \\
        w/o Regional Adaptive Attention & 0.871 & 0.023 & 0.929 & 0.16\% & 1.65 \\
        w/o Multi-Resolution DVF & 0.857 & 0.026 & 0.918 & 0.22\% & 1.63 \\
        w/o Both Attention Mechanisms & 0.843 & 0.029 & 0.912 & 0.25\% & 1.79 \\
        \bottomrule
    \end{tabular}
\end{table*}

The ablation studies presented in Table \ref{tab:ablation} confirm that each component of the proposed AD-RegNet architecture contributes significantly to overall registration performance. Removing the bidirectional cross-attention module results in a performance drop, with TRE increasing from 1.51 mm to 1.61 mm, a 0.06 mm (3.9\%). Excluding the regional adaptive attention results in a moderate performance decrease, with TRE rising to 1.65 mm, an increase of 0.10 mm (6.5\%). Eliminating the multi-resolution DVF synthesis also impacts accuracy, with TRE increasing to 1.63 mm, a 0.08 mm (5.2\%) degradation. Finally, removing both attention mechanisms yields the most significant drop in performance, with TRE increasing to 1.79 mm, representing an overall increase of 0.24 mm (15.5\%) compared to the full model.

\section{Discussion}
\label{sec:discussion}
Our experiments with the proposed AD-RegNet framework demonstrate several key findings. Attention mechanisms significantly improve registration accuracy, with the bidirectional cross-attention approach showing particular effectiveness in establishing correspondences between moving and fixed images. The regional adaptive attention mechanism enables focusing on anatomically relevant structures, resulting in improved registration accuracy in regions of interest. The multi-resolution deformation field synthesis is crucial for capturing both global and local deformations, especially in cases of large lung deformations. Additionally, the method achieves a favorable balance between registration accuracy and computational efficiency, making it suitable for clinical applications. The proposed approach generalizes well across different anatomical structures (brain and lung) and imaging modalities (MRI and CT), demonstrating its versatility. 

Despite these promising results, the method has several limitations that could be addressed in future work. The current implementation requires patch-based training for high-resolution 3D volumes due to memory constraints; more memory-efficient attention mechanisms or mixed-precision training could enable full-volume processing. More challenging multi-modal scenarios, such as MR-CT registration, were not explored, and future extensions could handle a wider range of modality combinations. Additionally, the current evaluation is limited to publicly available datasets; future work should include clinical validation on diverse patient populations and pathologies to assess the method's robustness in real-world clinical settings. 

\section{Conclusion}
\label{sec:conclusion}
In this work, we presented a novel Attention-Driven Framework for Non-Rigid Medical Image Registration (AD-RegNet). Our approach employs attention mechanisms to guide the registration process, incorporating a bidirectional cross-attention module to establish correspondences between moving and fixed images at multiple scales, a regional adaptive attention mechanism to focus on anatomically relevant structures, and a multi-resolution deformation field synthesis strategy for accurate alignment. Experimental results on two distinct datasets (DIRLab for thoracic CT and IXI for brain MRI) demonstrate the versatility and effectiveness of our approach across different anatomical structures and imaging modalities. AD-RegNet achieves competitive performance with state-of-the-art methods, with a Dice Similarity Coefficient of 0.759 $\pm$ 0.121 on the IXI dataset and a Target Registration Error of 1.51(1.39) mm on the DIRLab dataset, while maintaining a favorable balance between registration accuracy and computational efficiency.

Ablation studies confirm the importance of each component in our architecture, with bidirectional cross-attention and multi-resolution DVF synthesis providing the most significant contributions to performance. The comparison with different attention mechanisms highlights the advantages of our attention-driven framework, which achieves a better balance between accuracy and efficiency compared to self-attention-based methods. Future work will focus on addressing the limitations discussed, including memory constraints, multi-modal registration, real-time performance, and clinical validation. We believe that AD-RegNet represents a significant step forward in medical image registration and has the potential to improve various clinical applications, including disease diagnosis, treatment planning, and image-guided interventions.

\section*{Acknowledgment}
The authors would like to thank the creators and maintainers of the DIRLab and IXI datasets for making their data publicly available.

\bibliographystyle{elsarticle-num} 
\bibliography{biblo}

\end{document}